\documentclass[10pt,journal,compsoc]{IEEEtran}

\usepackage{graphicx}
\usepackage{amssymb}
\usepackage{ctable}
\usepackage{multirow}
\usepackage{makecell}
\usepackage{url}
\usepackage[T1]{fontenc}

\newcommand\MYhyperrefoptions{bookmarks=true,bookmarksnumbered=true,
pdfpagemode={UseOutlines},plainpages=false,pdfpagelabels=true,
colorlinks=true,linkcolor={black},citecolor={black},urlcolor={black},
pdftitle={Bare Demo of IEEEtran.cls for Computer Society Journals},
pdfsubject={Typesetting},
pdfauthor={Michael D. Shell},
pdfkeywords={Computer Society, IEEEtran, journal, LaTeX, paper,
             template}}

\usepackage[\MYhyperrefoptions,pdftex]{hyperref}

\ifCLASSOPTIONcompsoc
  \usepackage[nocompress]{cite}
\else
  \usepackage{cite}
\fi

\ifCLASSINFOpdf
\else
\fi

\usepackage{array}

\ifCLASSOPTIONcompsoc
  \usepackage[caption=false,font=footnotesize,labelfont=sf,textfont=sf]{subfig}
\else
  \usepackage[caption=false,font=footnotesize]{subfig}

\fi

\begin{document}

\title{Accurate, Data-Efficient, Unconstrained Text Recognition with Convolutional Neural Networks}

\author{Mohamed~Yousef, 
        Khaled~F.~Hussain,
        and~Usama~S.~Mohammed
\IEEEcompsocitemizethanks{\IEEEcompsocthanksitem Mohamed Yousef, 
and Khaled F. Hussain are with Computer Science Department, Faculty of
Computers and Information, Assiut University, Asyut 71515, Egypt  (e-mail: mohamed.mahdi@fci.au.edu.eg ; khussain@aun.edu.eg).\protect\\
\IEEEcompsocthanksitem Usama S. Mohammed is with Electrical Engineering Department, Faculty of
Engineering, Assiut University, Asyut 71515, Egypt (e-mail: usama@aun.edu.eg).
}}

\markboth{Journal of \LaTeX\ Class Files,~Vol.~14, No.~8, August~2015}%
{Shell \MakeLowercase{\textit{et al.}}: Bare Demo of IEEEtran.cls for Computer Society Journals}

\IEEEtitleabstractindextext{%
\begin{abstract}
Unconstrained text recognition is an important computer vision task, featuring a wide variety of different sub-tasks, each with its own set of challenges. One of the biggest promises of deep neural networks has been the convergence and automation of feature extractors from input raw signals, allowing for the highest possible performance with minimum required domain knowledge. To this end, we propose a data-efficient, end-to-end neural network model for generic, unconstrained text recognition. In our proposed architecture we strive for simplicity and efficiency without sacrificing recognition accuracy. Our proposed architecture is a fully convolutional network without any recurrent connections trained with the CTC loss function. Thus it operates on arbitrary input sizes and produces strings of arbitrary length in a very efficient and parallelizable manner. We show the generality and superiority of our proposed text recognition architecture by achieving state of the art results on seven public benchmark datasets, covering a wide spectrum of text recognition tasks, namely: Handwriting Recognition, CAPTCHA recognition, OCR, License Plate Recognition, and Scene Text Recognition. Our proposed architecture has won the ICFHR2018 Competition on Automated Text Recognition on a READ Dataset.
\end{abstract}

}

\maketitle

\IEEEdisplaynontitleabstractindextext

\IEEEpeerreviewmaketitle

\IEEEraisesectionheading{\section{Introduction}\label{sec:introduction}}

\IEEEPARstart{T}{ext} recognition is considered one of the earliest computer vision tasks to be tackled by researches. For more than a century now \cite{schantz1982history} since its inception as a field of research, researchers never stopped working on it. This can be contributed to two important factors. 

First, the pervasiveness of text and its importance to our everyday life, as a visual encoding of language that is used extensively in communicating and preserving all kinds of human thought. Second, the necessity of text to humans and its pervasiveness has led to big adequacy requirements over its delivery and reception which has led to the large variability and ever-increasing visual forms text can appear in. Text can originate either as printed or handwritten, with large possible variability in the handwriting styles, the printing fonts, and the formatting options. Text can be found organized in documents as lines, tables, forms, or cluttered in natural scenes. Text can suffer countless types and degrees of degradations, viewing distortions, occlusions, backgrounds, spacings, slantings, and curvatures. Text from spoken languages alone (as an important subset of human-produced text) is available in dozens of scripts that correspond to thousands of languages. All this has contributed to the long-standing complicated nature of unconstrained text recognition.

Since a deep Convolutional Neural Network (CNN) \cite{fukushima1982neocognitron,lecun1998gradient} won the ImageNet image classification challenge \cite{krizhevsky2012imagenet}, Deep Learning based techniques have invaded most tasks related to computer vision, either equating or surpassing all previous methods, at a fraction of the required domain knowledge and field expertise. Text recognition was no exception, and methods based on CNNs and Recurrent Neural Networks (RNNs) have dominated all text recognition tasks like OCR \cite{breuel2013high}, Handwriting recognition \cite{graves2009novel}, scene text recognition \cite{jaderberg2016reading}, and license plate recognition \cite{li2016reading}, and have became the de facto standard for the task. 

Despite their success, one can spot a number of shortcomings in these works. First, for many tasks, an RNN is required for achieving state of the art results which brings-in non-trivial latencies due to the sequential processing nature of RNNs. This is surprising, given the fact that, for pure-visual text recognition long range dependencies have no effect and only local neighborhood should affect the final frame or character classification results. Second, for each of these tasks, we have a separate model that can, with its own set of tweaks and tricks, achieve state of the art result in a single task or a small number of tasks. Thus, No single model is demonstrated effective on the wide spectrum of text recognition tasks. Choosing a different model or feature-extractor for different input data even inside a limited problem like text recognition is a great burden for practitioners and clearly contradicts the idea of automatic or data-driven representation learning promised by deep learning methods.

In this work, we propose a very simple and novel neural network architecture for generic, unconstrained text recognition. Our proposed architecture is a fully convolutional CNN \cite{long2015fully} that consists mostly of depthwise separable convolutions \cite{sifre2014rigid} with novel inter-layer residual connections \cite{he2016deep} and gating, trained on full line or word labels using the CTC loss \cite{graves2006connectionist}. We also propose a set of generic data augmentation techniques that are suitable for any text recognition task and show how they affect the performance of the system. We demonstrate the superior performance of our proposed system through extensive experimentation on seven public benchmark datasets. We were also able, for the first time, to demonstrate human-level performance on the reCAPTCHA dataset proposed recently in a Science paper \cite{george2017generative}, which is more than 20\% absolute increase in CAPTCHA recognition rate compared to their proposed RCN system. We also achieve state of the art performance in SVHN \cite{netzer2011reading} (the full sequence version), the unconstrained settings of IAM English offline handwriting dataset \cite{marti2002iam}, KHATT Arabic offline handwriting dataset  \cite{mahmoud2014khatt}, University of Washington (UW3) OCR dataset \cite{phillips1996user},  AOLP license plate recognition dataset \cite{hsu2013application} (in all divisions). Our proposed system has also won the ICFHR2018 Competition on Automated Text Recognition on a READ Dataset \cite{strauss2018icfhr2018} achieving more than 25\% relative decrease in Character Error Rate (CER) compared to the  entry achieving the second place.

To summarize, we address the unconstrained text recognition problem. In particular, we make the following contributions:
\begin{itemize}
\item We propose a simple, novel neural network architecture that is able to achieve state of the art performance, with feed forward connections only (no recurrent connections), and using only the highly efficient convolutional primitives.
\item We propose a set of data augmentation procedures that are generic to any text recognition task and can boost the performance of any neural network architecture on text recognition.
\item We conduct an extensive set of experiments on seven benchmark datasets to demonstrate the validity of our claims about the generality of our proposed architecture. We also perform an extensive ablation study on our proposed model to demonstrate the importance of each of its submodules.
\end{itemize}

Section \ref{sec:rw} gives an overview of related work on the field. Section \ref{sec:meth} describes our architecture design and its training process in details. Section \ref{sec:exp} describes our extensive set of experiments and presents its results.

\begin{figure*}[h!]
	\centering
	\subfloat[Overall proposed architecture]{\includegraphics[width=0.4\textwidth]{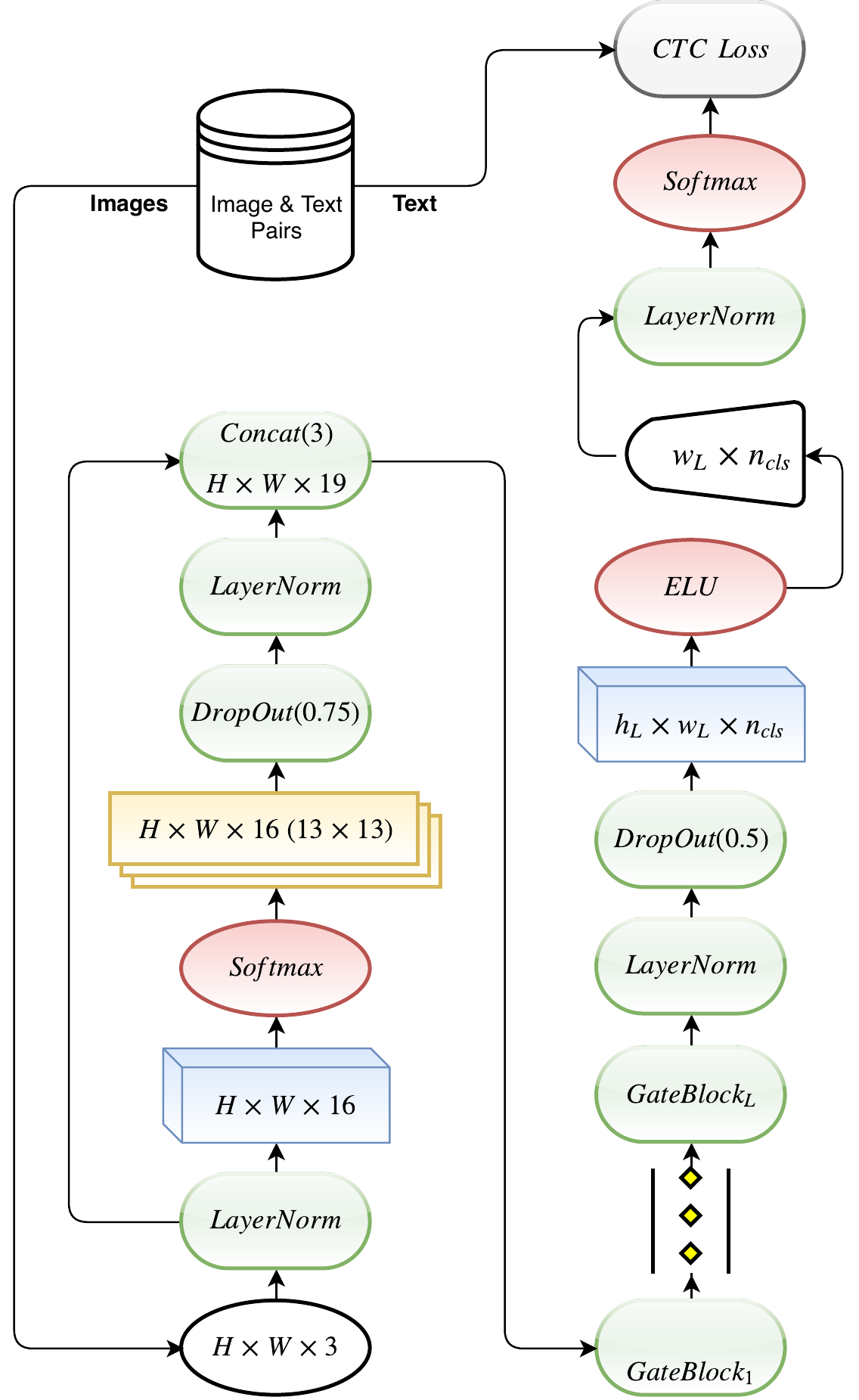}}
	\subfloat[Details of the repeated $GateBlock_i$]{\includegraphics[width=0.4\textwidth]{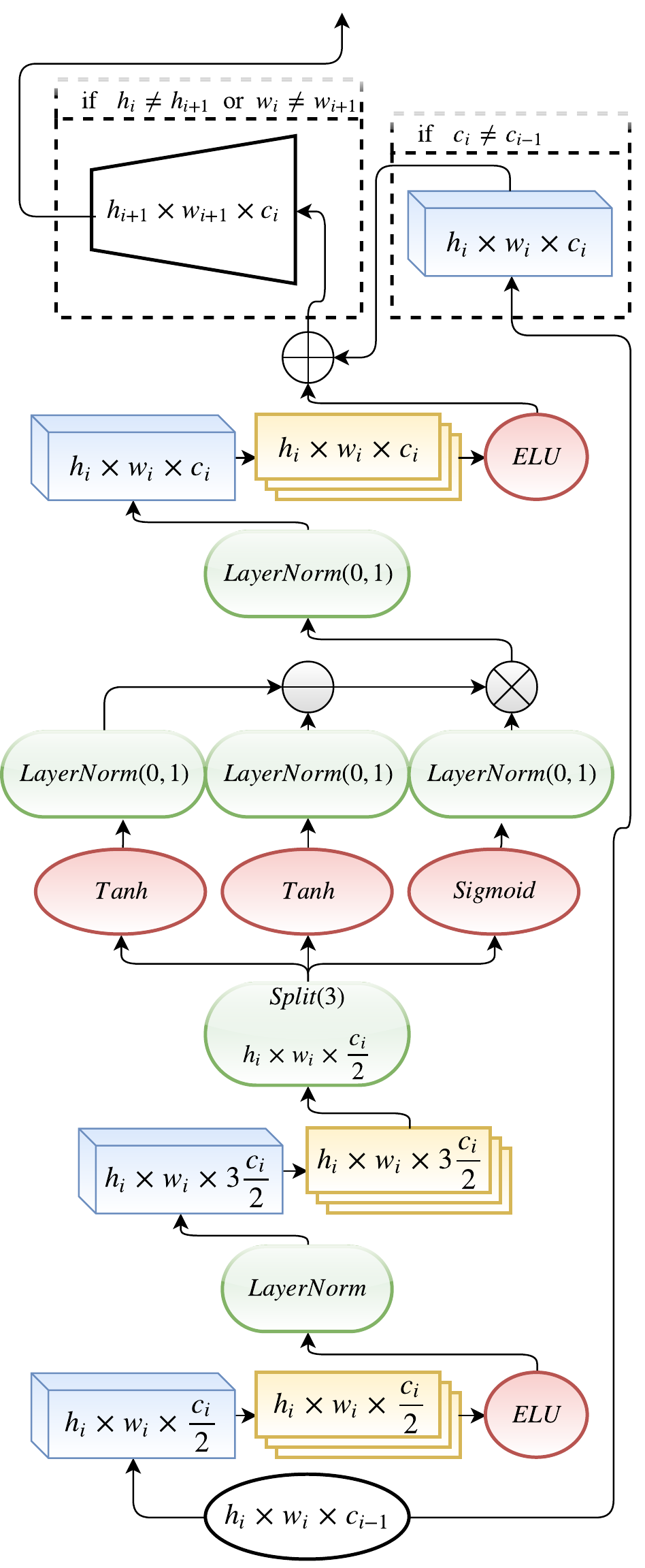}} \\
	\subfloat{\includegraphics[width=0.8\textwidth]{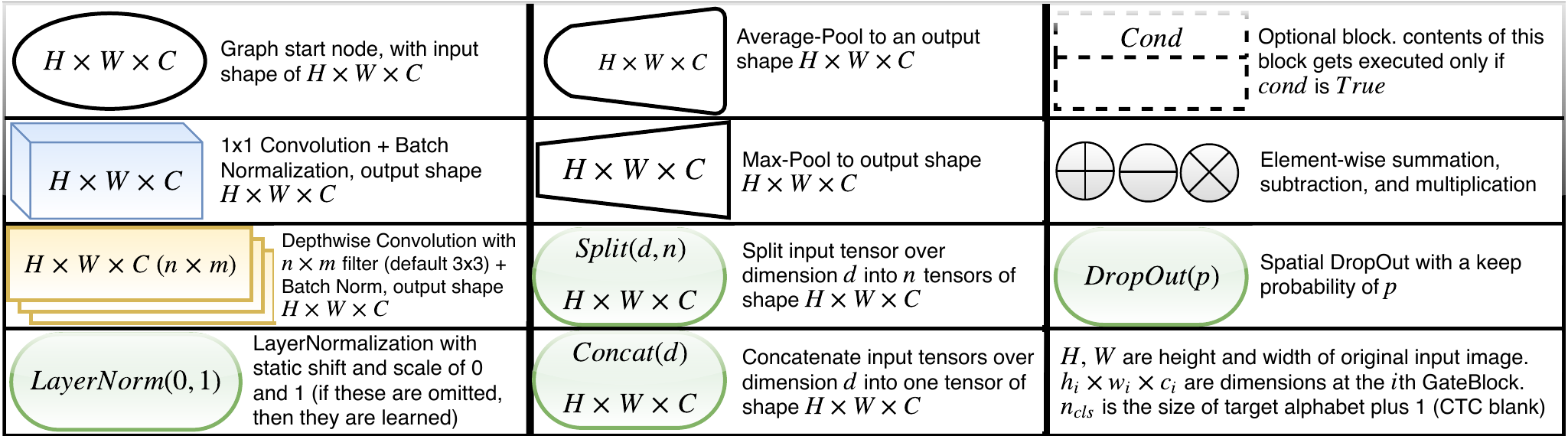}}
    
     \caption{Detailed diagram of our proposed network. We explicitly present details of the network, with clear description of each block in the legend below the diagram. We also highlight each dimensionality change of input tensors as they flow through the network. (a) Our overall architecture. Here, we assume the network has $L$ stacked layers of the repeated $GateBlock$. (b) Level $i$ of the repeated $GateBlock$.}\label{fig:network}
    
\end{figure*}

\section{Related Work}
\label{sec:rw}
Work in text recognition is enormous and spans over a century, we here focus on methods based on deep learning as they have been the state of the art for at least five years now. Traditional methods are reviewed in \cite{zhu2016scene,ye2015text}, and very old methods are reviewed in \cite{schantz1982history}.

There is two major trends in deep learning based sequence transduction problems, both avoid the need for fine-grained alignment between source and target sequences. The first is using CTC \cite{graves2006connectionist}, and the second is using sequence-to-sequence (Encoder-Decoder) models \cite{cho2014learning,sutskever2014sequence} usually with attention \cite{bahdanau2014neural}.

\subsection{CTC Models}
In these models the neural network is used as an emission model, such that the input signal is divided into frames and the model computes, for each input frame, the likelihood of each symbol of the target sequence alphabet. A CTC layer then computes, for a given input sequence, a probability distribution over all possible output sequences. An output sequence can then be predicted either greedily (i.e. taking the most likely output at each time-step) or using beam-search. A thorough explanation of CTC can be found in \cite{hannun2017sequence}.

BLSTM \cite{graves2005framewise} + CTC was used to transcribe online handwriting data on \cite{graves2008unconstrained}, they were also used for OCR in \cite{breuel2013high}. Deep BLSTM + CTC were used for handwriting recognition (with careful preprocessing) in \cite{bluche2015deep}, they were also used on HOG features for scene text recognition in \cite{su2014accurate}. This architecture was mostly abandoned in favor of CNN+BLSTM+CTC.

Deep MDLSTM \cite{graves2007multi} (interleaved with convolutions) + CTC have achieved the best results on offline handwriting recognition since their introduction in \cite{graves2009offline} until recently \cite{bluche2014a2ia,pham2014dropout,voigtlaender2016handwriting}. However, their performance was recently equated with the much faster CNN+BLSTM+CTC architecture \cite{puigcerver2017multidimensional,duttaimproving}. MDLSTMs have also shown good results in speech recognition \cite{li2016exploring}.

Various flavors of CNN+BLSTM+CTC currently deliver state of the results in many text recognition tasks. Breuel \cite{breuel2017high} shows how they improve over BLSTM+CTC for OCR tasks. Li \emph{et al.} \cite{li2016reading} shows state of the art license plate recognition using them. Shi \emph{et al.} \cite{shi2017end} uses it for scene text recognition, however, it is not currently the state of the art. Puigcerver \cite{puigcerver2017multidimensional} and Dutta \emph{et al.}  \cite{duttaimproving} used CNN+BLSTM+CTC to equate the performance of MDLSTMs in offline handwriting recognition.

There is a recent shift to recurrence-free architectures in all sequence modeling tasks \cite{bai2018empirical} due to their much better parallelization performance. We can see the trend of CNN+CTC in sequence transduction problems as used in \cite{wang2017residual} for speech recognition, or in \cite{gao2017reading,yin2017scene} for scene text recognition. The main difference between these later works and ours is that while we also utilize a CNN+CTC (1) we use a completely different CNN architecture (2) while previous CNN+CTC work was mostly demonstrative of the viability/competitiveness of the approach and was carried on a specific task, we show strong state of the art results on a wide range of text recognition tasks.

\subsection{Encoder-Decoder Models}
These models consist of two main parts. First, the encoder reads the whole input sequence, then the decoder produces the target sequence token-by-token using information produced by the decoder. An attention mechanism is usually utilized to allow the decoder to automatically (soft-)search for parts of the source sequence that are relevant to predicting the current target token \cite{cho2014learning,sutskever2014sequence,bahdanau2014neural}.

In \cite{shi2018aster} a CNN+BLSTM is used as an encoder and a GRU \cite{cho2014learning} is used as a decoder for scene text recognition. A CNN encoder and an RNN decoder is used for reading street name signs from multiple views in \cite{wojna2017attention}. Recurrent convolutional layers \cite{liang2015recurrent} are used as an encoder and an RNN as a decoder for scene text recognition in \cite{lee2016recursive}. In \cite{chowdhury2018efficient} a CNN+BLSTM is used as an encoder and an LSTM is used as a decoder for handwriting recognition. They achieve competitive results, but their inference time is prohibitive, as they had to use beam search with very wide beam to achieve a good result. Bluche \emph{et al.} \cite{bluche2017scan} tackles full paragraph recognition (without line segmentation) they use a MDLSTM as an encoder, LSTM as a decoder, and another MDLSTM for computing attention weights.

\section{Methodology}
\label{sec:meth}
In this section, we present details of our proposed architecture. First, Section \ref{sub:over} gives an overview of the proposed architecture. Then, Section \ref{sub:gbs} presents derivation of the gating mechanism utilized by our architecture. After that, Section \ref{sub:da} discusses the data augmentation techniques we use in the course of training. Finally, Section \ref{sub:mt} describes our model training and evaluation process.

\subsection{Overall Architecture}
\label{sub:over}
Figure \ref{fig:network} presents a detailed diagram of the proposed architecture. We opted to use a very detailed diagram for illustration instead of a generic one, in order to make it straightforward to implement the proposed architecture in any of the popular deep learning frameworks without much guessing.

As shown in Figure \ref{fig:network} (a), our proposed architecture is a fully convolutional network (FCN) that makes heavy use of both Batch Normalization \cite{ioffe2015batch} and Layer Normalization \cite{ba2016layer} to both increase convergence speed and regularize the training process. We also use Batch Renormalization \cite{ioffe2017batch} on all Batch Normalization layers. This allows us to use batch sizes as low as 2 with minor effects on final classification performance. Our main computational block is Depthwise separable convolutions. Using them instead of regular convolutions gives the same or better classification performance, but at a much faster convergence speed and a drastic reduction in parameter count and computational requirements. We also found the inductive bias of depthwise convolutions to have a strong regularization effect on training. For example, for our original architecture with regular convolutions, we had to use spatial DropOut \cite{tompson2015efficient} (i.e. randomly dropping entire channels completely) after nearly every convolution, in contrast to a single spatial DropOut layer after the stack of $GateBlock$s as we currently do. An important point to note here, is that in contrast to other works that use stacks of depthwise separable convolutions \cite{chollet2017xception,kaiser2017depthwise}. We found it crucial to use Batch Normalization in-between the depthwise convolution and the 1x1 pointwise convolution \cite{lin2013network}. We also note here that we found spatial DropOut to provide much better regularization than regular un-structured DropOut \cite{srivastava2014dropout} on the whole tensor.

First, the input image is Layer Normalized, input features are then projected to be 16 using a 1x1 convolution, then the input tensor is softmax-normalized as this was found to greatly increase convergence speed on most datasets. Next, each channel is preprocessed independently with a 13x13 filter using a depthwise convolution, then we employ the only used dense connection by concatenating, on the channel dimension, the result of preprocessing with the layer-normalized original image. The result is fed to a stack of $GateBlock$s that are responsible for the majority of the input sequence modeling, and target sequence generation task through a fixed number of in-place whole-sequence transformation stages. This was demonstrated to be effectively an unrolled iterative estimation \cite{greff2016highway} of target sequence. Spatial DropOut is applied on the output of the last $GateBlock$, then the channel dimension is projected to be equal to the size of the target alphabet plus one (required for the CTC blank \cite{graves2006connectionist}) using a 1x1 convolution. Finally, we perform global average pooling on the height dimension. This step enables the network to handle inputs of any height. The result is then layer normalized, softmax-normalized then fed to the CTC loss along with its corresponding text label during training.

\subsection{$GateBlock$s}
\label{sub:gbs}
Stacked $GateBlock$s are the main computational blocks of our model. Using attention gates to control the inter-layer Information flow, in the sense of filtering-out unimportant signals (e.g.  background noise) and sharpening/intensifying important signals, has received a lot of interest by many researchers lately \cite{wangfei2017residual,hu2017squeeze}. However, the original concept is old \cite{hochreiter1997long,jacobs1991adaptive}. Many patterns of gating mechanisms has been proposed in literature, first for connecting RNNs time steps as a technique to mitigate the vanishing gradient problem \cite{hochreiter1997long,gers2000learning}, then this idea was extended to connecting layers of any feed-forward neural networks. This was done either for training very deep networks \cite{srivastava2015highway,kalchbrenner2015grid} or for using it, as stated previously, as a smart, content-aware filters.

However, most of the attention gates that were suggested recently for connecting CNN layers use ad-hoc mechanisms tested only on the target datasets without comparison to other mechanisms. We opted to base our attention gates on the gating mechanism suggested by Highway Networks \cite{srivastava2015highway}, as they are based on the widely used, heavily studied LSTM networks. Also, a number of variants of Highway networks were studied, compared and contrasted in subsequent studies \cite{kim2016character,greff2016highway}. We also validate the superiority of these gates experimentally in Section \ref{sec:exp}.

Assume a plain feed-forward $L$-layer neural network, with layers $y_1 \dots y_L$. Each of these layers applies a non-linear transformation $H$ to its input to get its output, such that
\begin{equation}
y_{i+1} = H(y_i)
\end{equation}
In a Highway network, we have two additional non-linear transformations $T$ and $C$. These act as gates that can dynamically pass part of its inputs and suppress the other part, conditioned on the input itself. Therefore:
\begin{equation}
y_{i+1} = H(y_i) \cdot T(y_i) + y_i \cdot C(y_i)
\end{equation}
Authors suggest to tie the two gates, such that we have only one gate $T$ and the other is its opposite $C = 1-T$. This way we find out
\begin{equation}
y_{i+1} = H(y_i) \cdot T(y_i) + y_i \cdot [1-T(y_i)]
\end{equation}
This can re-factored to 
\begin{equation}
y_{i+1} = \Big[ \big(H(y_i) - y_i\big) \cdot T(y_i) \Big] + y_i
\label{eqn:hres}
\end{equation}
We should note here that the dimensionality of $y_i$, $y_{i+1}$, $H(y_i)$, and $T(y_i)$ must be the same, for this equation to be valid. This is can be an important limitation from both the computational and memory usage perspectives for wide networks (e.g for CNNs when each of those tensors has a large number of channels) or even small networks when $y_i$ have big spatial dimensions.

If we would use the same original plain network but with residual connections we would have
\begin{equation}
y_{i+1} = \Big[ H(y_i) \Big] + y_i
\label{eqn:res}
\end{equation}
Comparing Equations \ref{eqn:hres} and \ref{eqn:res}, it is evident that both equations have the same residual connection in the right hand-side of the equation and thus we can interpret the left hand-side of Equation \ref{eqn:hres} as the transformation function applied to input $y_i$. This interpretation motivates us to perform two modifications to Equation \ref{eqn:hres}.

First, we note that the left hand-side of Equation \ref{eqn:hres} performs a two-stage filtering operation on its input $y_i$ before adding the residual connection. The output of the transformation function $H$ is first subtracted from the input signal $y_i$, then multiplied by the gating function $T$. We argue that all transformations to the input signal should be learned and should be conditioned on the input signal $y_i$. So instead of using a single transformation function $H$, we propose using two transformation functions $H_1$ and $H_2$, such that:
\begin{equation}
y_{i+1} = \Big[ \big(H_1(y_i) - H_2(y_i) \big) \cdot T(y_i) \Big] + y_i
\label{eqn:hres2}
\end{equation}
In all experiments in this paper, $H_1$, $H_2$, and $T$ are implemented as Depthwise separable convolutions with $tanh$, $tanh$, and $sigmoid$ activation functions respectively.

Second, the residual connection interpretation along with the dimensionality problem we noted above, motivates us to make a simple modification to Equation \ref{eqn:hres}. This would allow us to efficiently utilize Highway gates even for wide and deep networks.

Let $P_1$ be a transformation mapping $x \in \mathbb{R}^{H \times W \times C}$ to $x' \in \mathbb{R}^{H' \times W' \times C'}$ and $P_2$ the opposite transformation; mapping $x' \in \mathbb{R}^{H' \times W' \times C'}$ to $x \in \mathbb{R}^{H \times W \times C}$. We can change Equation \ref{eqn:hres} such that
\begin{equation}
y'_i = P1(y_i)
\end{equation}
\begin{equation}
y_{i+1} = P2 \Big( \Big[ \big(H_1(y'_i) - H_2(y'_i) \big) \cdot T(y'_i) \Big] \Big) + y_i
\label{eqn:ctres}
\end{equation}
When using a sub-sampling transformation for $P_1$ this allows us to retain optimization benefit of residual connections (right hand-side of Equation \ref{eqn:ctres}) whilst computing Highway gates on a lower dimensional representation of $y_i$ which can be performed much faster and uses less memory. In all our experiments in this paper, $P_1$ and $P_2$ are implemented as Depthwise separable convolutions with the Exponential Linear Unit (ELU) \cite{clevert2015fast} as an activation function. Also, we always set $C'=\frac{C}{2}$, $H'=H$, and $W'=W$. This means that sub-sampling is only done on the channel dimension and spatial dimensions are kept the same.

\subsection{Data Augmentation}
\label{sub:da}
Encoding domain knowledge into data through label-preserving data transformations has always been very successful and had provided fundamental contributions to the success of data-driven learning systems in the past, and lately after the recent revival of neural networks. We here describe a number data augmentation techniques that are generic and can benefit most text recognition problems, these techniques are applied independently to any input data sample during training

\subsubsection{Projective Transforms}
Text lines or words to be recognized can suffer various types of scaling or shearing and more generally projection transforms. We apply a random projection transform to every input training sample, by randomly sampling four points that correspond to the original four corners of the original image (i.e. the sampled points represent the new corners of the image) under the following criteria 
\begin{itemize}
\item We either change the $x$ coordinate of all four points and fix the $y$ coordinate or vice versa. We found it almost always better to either change the image vertically or horizontally but not both simultaneously.
\item When changing either vertically or horizontally, the new distance between any connected pair of corners must not be less than half the original one, or more than double the original one.
\end{itemize}

\subsubsection{Elastic Distortions}
Elastic distortions were first proposed by \cite{simard2003} as a way for emulating uncontrolled oscillations of hand muscles. It was then used in \cite{Ciregan2012} for augmenting images of handwritten digits. In \cite{simard2003}, the authors apply a displacement field to each pixel in the original image, such that each pixel moves by a random amount in both the $x$ and $y$ directions. The displacement field is built by convolving a randomly initialized field with a Gaussian kernel. An obvious problem with this technique is that it operates at a very fine level (the pixel level), which makes it more likely that the resulting image is unrealistically distorted.

To counter this problem, we take the approach used by \cite{wigington2017data} and \cite{bloice2017augmentor}, by working at a coarser level to better preserve the localities of the input image. We generate a regular grid and apply a random displacement field to the control points of the grid. The original image is then warped according to the displaced control points. It is worth noting that the use of a coarse grid makes it unnecessary to apply a smoothing Gaussian to the displacement filed. However, unlike \cite{wigington2017data} we do not align to the baseline of the input image, which makes it much simpler to implement. We also impose the following criteria on the randomly generated displacement field, which is uniformly sampled in our implementation:
\begin{itemize}
\item We apply the distortions either to the $x$ direction or the $y$ direction and not to both. As we did for the projective transforms, we also found here it is better to apply distortions to one dimension at a time.
\item We force the minimum width or height of a distorted grid rectangle to be $1$. In other words, zero or negative values, although possible, are not allowed. 
\end{itemize}

\subsubsection{Sign Flipping}
We randomly flip the sign of the input image tensor. This is motivated by the fact that text content is invariant to its color or texture, and one of the simplest methods to alarm this invariance is sign flipping. It is important to stress that this augmentation improves performance even if all training and testing images have the same text and background colors. The fact that color-inversion data augmentation improves text recognition performance was previously noticed in \cite{cubuk2018autoaugment}.

\subsection{Model Training and Evaluation}
\label{sub:mt}
We train our model using the CTC loss function. This enables using unsegmented pairs of line-images and corresponding text transcriptions to train the model without any character/frame-level alignment. Every epoch, training instances are sampled without replacement from the set of training examples. Previously described augmentation techniques are applied during training on a batch-wise manner, i.e. the same random generated values are applied to the whole batch.

In order to generate target sequences, we simply use CTC greedy decoding \cite{graves2006connectionist}: i.e. taking the most probable label at each time step, then mapping the result using the CTC sequence mapping function which first removes repeated labels, then removes the blanks. In all our experiments, we do not utilize any form of language modeling or any other non-visual cues.

We evaluate the quality of our models based on the popular Levenstein edit distance \cite{levenshtein1966binary} evaluated on the character level, and normalized by the length of the ground truth, which is commonly known as Character Error Rate (CER). Due to the inherent ambiguity in some of the datasets used for evaluation (e.g. handwriting datasets), we need another metric to illustrate how much of this ambiguities have been learned by our model, and by how much can the model performance increase if another signal (other than the visual one, e.g. a linguistic one) can help ranking the possible output sequence transcriptions. For this task, we use the CER@Top$N$ metric, which computes the CER as if we were allowed to choose for each output sequence the candidate with the lowest CER out of those returned from CTC beam search decoder \cite{graves2006connectionist}. This gives us an upper bound on the maximum possible performance gain by using a secondary re-ranking signal.

We use Adam \cite{kingma2014adam} to optimize our network parameters, also the base learning rate is exponentially decayed during the training process. We also evaluate our models using Polyak averaging \cite{polyak1992acceleration} at inference time.

\section{Experiments}
\label{sec:exp}
In this section we evaluate and compare the proposed architecture to the previous state of the art techniques by conducting extensive experiments on seven public benchmark datasets covering a wide spectrum of text recognition tasks.

\subsection{Implementation Details}
\label{sub:is}
For all our experiments, we use an initial learning rate of $5 \times 10^{-3}$, which is exponentially decayed such that it reaches $0.1$ of its value after $9 \times 10^4$ batches. We use the maximum batch allowed by our platform with a minimum size of $2$. All our experiments are performed on a machine equipped with a single Nvidia TITAN Xp GPU. All our models are implemented using TensorFlow \cite{abadi2016tensorflow}.

The variable part of our model is the $GateBlock$ stack, which is (in our experiments) parameterized by three parameters written as $n(c_1,c_2)$. $n$ is the number of $GateBlock$s, $c_1$ and $c_2$ are the number of channels in the first and third $GateBlock$. For all instances of our model used in the experiments, the number of channels is changed only in the first and third $GateBlock$. 

\subsection{Vicarious reCAPTCHA Dataset}
\begin{table}[h!]
\caption{Results of our method compared to state of the art on the Vicarious reCAPTCHA Dataset. \textbf{SER} is the whole sequence error rate, \textbf{Aug.} is whether data augmentation is applied (True) or not (False). \textbf{Case} is whether error metrics are case sensitive (True) or not (False). Best results are in bold }
\label{table:vrec}
\centering
\begin{tabular}{c c c c c c} 
 \specialrule{.3em}{.2em}{.2em}
 Method & Input Scale & SER(\%) & CER(\%) & Aug. & Case \\
 \specialrule{.3em}{.2em}{.2em}
 RCN   \cite{george2017generative}   & 2$X$  & 33.4 & 5.7 & True & False \\ 
 Human \cite{george2017generative}   & 1$X$  & 12.6 & - & - & False \\
 Ours: 8(64,512)  & 1$X$  & 21.70 & 3.77 & False & True \\
 Ours: 8(64,512)  & 1$X$  & 12.86 & 2.14 & True  & True \\
 Ours: 16(64,512) & 1$X$  & \textbf{12.30} & \textbf{2.04} & True & True \\
 \specialrule{.3em}{.2em}{.2em}
\end{tabular}
\end{table}

In a recent Science paper \cite{george2017generative}, Vicarious presented RCN, a system that could break most modern text-based CAPTCHAs in a very data-efficient manner, they claimed it is multiple orders of magnitude more data efficient than rivaling deep-learning based systems in breaking CAPTCHAs and also more accurate. To support those claims they collected a number of datasets from multiple online CAPTCHA systems and annotated them using humans. We choose to compare against the reCAPTCHA subset specifically since they have also done an analysis of human performance on this specific subset only (shown in Table \ref{table:vrec}). The reCAPTCHA dataset consists of 5500 images, 500 for training and 5000 for testing, with the critical aspect being generalizing to unseen CAPTCHAs using only the 500 training images. In our experiments, we split the training images to 475 images used for training and 25 images used for validation.

Before comparing our results directly to RCN's, it is important to note that all RCN's evaluations are case-insensitive, while our evaluations use the much harder case-sensitive evaluation, the reason we did so is that case-sensitive evaluation is the norm for text recognition and we were already getting accurate results without problem simplification.

As shown in Table \ref{table:vrec}, without data augmentation, and with half input spatial resolution our system is already 10\% more accurate than RCN. In fact, it alot more accurate since RCN's results are case-insensitive. When we apply data augmentations and use a deeper model, our model is 20\% better than RCN, and more importantly, it is able to surpass the human performance estimated in their paper.

\subsection{SVHN}
\begin{table}[h!]
\caption{Results of our method compared to state of the art on SVHN. \textbf{SER} is the whole sequence error rate. Parameters are in millions. Best results are in bold }
\label{table:svhn}
\centering
\begin{tabular}{c c c c c} 
 \specialrule{.3em}{.2em}{.2em}
 Method & Input Scale & SER(\%) & Params & Year \\
 \specialrule{.3em}{.2em}{.2em}
 11 layer Maxout CNN \cite{goodfellow2013multi}   & 64x64  & 3.96 & 51 & 2013 \\ 
 Single DRAM \cite{ba2014multiple}                & 64x64  & 5.1  & 14 & 2014 \\
 Single DRAM MC \cite{ba2014multiple}             & 64x64  & 4.4  & 14 & 2014 \\
 Double DRAM MC \cite{ba2014multiple}             & 64x64  & 3.9  & 28 & 2014 \\
 ST-CNN Single \cite{jaderberg2015spatial}        & 64x64  & 3.7  & 33 & 2015 \\
 ST-CNN Multi \cite{jaderberg2015spatial}         & 64x64  & 3.6  & 37 & 2015 \\
 EDRAM Single \cite{ablavatski2017enriched}       & 64x64  & 4.36 & 11 & 2017 \\
 EDRAM Double \cite{ablavatski2017enriched}       & 64x64  & 3.6  & 22 & 2017 \\
 Ours: 4(128,512)   & 32x32  & 3.3  & 0.9 & 2018 \\ 
 Ours: 4(128,1024)  & 32x32  & \textbf{3.1}  & 3.4 & 2018 \\ 
 \specialrule{.3em}{.2em}{.2em}
\end{tabular}
\end{table}

Street View House Numbers (SVHN) \cite{netzer2011reading} is a challenging real-world dataset released by Google. This dataset contains around 249k real world images of house numbers divided to 235k images for training and validation and 13k for testing, the task is to recognize the sequence of numbers in each image. There are between 1 and 5 digits in each image, with a large variability in texture, scale, and spatial arrangement.

Following \cite{goodfellow2013multi} we formed a validation set of 5k images by randomly sampling images from the training set. We also convert input RGB images to grayscale, as it was noticed by many authors \cite{goodfellow2013multi,ba2014multiple} that this did not affect performance. Although most literature resizes the cropped digits image to 64x64 pixels, we found that for our model, 32x32 pixels was sufficient. This leads to much faster processing of input images. We only applied the data augmentations introduced in \ref{sub:da}.

As shown in Table \ref{table:svhn}, after progress in this benchmark has stalled since 2015 on a full sequence error of 3.6\%, we were able to advance this 3.3\% using only 0.9M parameters. That is a massive $25X$ reduction in parameter count from second best performing system and at a better classification accuracy. Using a larger model with 3.4M parameters we are able to advance the full sequence error to 3.1\%. 

\subsection{University of Washington Database III}
\begin{table}[h!]
\caption{Results of our method compared to state of the art on the University of Washington Database III. \textbf{Norm.} is whether text line geometric normalization is applied (True) or not (False). \textbf{Epochs} is the number of epochs of training the network took to converge. Best results are in bold }
\label{table:uw3}
\centering
\begin{tabular}{c c c c c} 
 \specialrule{.3em}{.2em}{.2em}
 Method & Input Scale & CER(\%) & Norm. & Epochs \\
 \specialrule{.3em}{.2em}{.2em}
 LSTM \cite{breuel2013high}   & 32 x W  & 0.60 & True & 100 \\ 
 LSTM \cite{breuel2017high}   & 48 x W  & 0.40 & True & 500 \\ 
 CNN-LSTM \cite{breuel2017high}      & 48 x W  & 0.17 & True & 500 \\ 
 CNN-LSTM \cite{breuel2017high}      & 48 x W  & 0.43 & False & 500 \\
 Ours: 4(128,512) & 32 x W  & 0.10 & False & 5 \\
 Ours: 8(128,512) & 32 x W  & \textbf{0.08} & False & 8 \\
 \specialrule{.3em}{.2em}{.2em}
\end{tabular}
\end{table}

The University of Washington Database III (UW3) dataset \cite{phillips1996user}, contains 1600 pages of document images from scientific journals and other common sources. We use text lines extracted by \cite{breuel2013high} which contains 100k lines split to 90k for training and 10k for testing. As an OCR benchmark it presents different sets of challenges like small original input resolutions, and the need for extremely low CER.

Geometric Line Normalization is introduced in \cite{breuel2013high} to aid text line recognition using 1D LSTM which are not translationally invariant along the vertical axis. It was also found essential for CNN-LSTM networks \cite{breuel2017high}. It was mainly modeled for targeting Latin scripts, which have many character pairs that are nearly identical in shape but are distinguished from each other based on their position and size relative to text-line's baseline and x-height. It is a fairly-complex, script specific operation.

The only preprocessing we perform on images is to resize it such that the height of the input text line is 32 pixels, while maintaining the aspect ratio of the line image.

As shown in Table \ref{table:uw3}, even a fairly small model without any form of input text line normalization or script specific knowledge is able to get state of the art CER of 0.10\% on the dataset. Using a deeper model, we are able to get a slightly better accuracy of 0.08\% CER. One important thing to also note from the table is that our models take two orders of magnitude less epochs to converge to the final classification accuracy, and use smaller input spatial dimensions.

\subsection{Application-Oriented License Plate (AOLP) Dataset}
\begin{table}[h!]
\caption{Results of our method compared to state of the art on the AOLP dataset. \textbf{S} is the whole sequence recognition accuracy. \textbf{C} is the character recognition accuracy. AC, LE, and RP are divisions of the AOLP dataset. Best results are in bold }
\label{table:aolp}
\centering
\begin{tabular}{c|c|c|c|c|c|c}
 \specialrule{.3em}{.2em}{.2em}
 \multirow{2}{*}{Method} & \multicolumn{2}{c}{AC} & \multicolumn{2}{c}{LE} & \multicolumn{2}{c}{RP} \\
 & S(\%) & C(\%) & S(\%) & C(\%) & S(\%) & C(\%) \\
 \specialrule{.3em}{.2em}{.2em}
 LBP+LDA\cite{hsu2013application}        & 88.5  &  96   & 86.6  & 94    & 85.7  & 95    \\ 
 CNN+BLSTM\cite{li2016reading}             & 94.85 &  -    & 94.19 & -     & 88.38 & -     \\
 DenseNet \cite{wu2018many}                & 96.61 & 99.08 & \textbf{97.80} & \textbf{99.65} & 91.00 & 97.22 \\
 Ours: 4(128,512)& \textbf{97.63} & \textbf{99.56} & 97.65 & 98.95 & 92.93 & 98.27 \\
 Ours: 8(128,512)& \textbf{97.63} & \textbf{99.56} & 97.65 & 98.95 & \textbf{94.57} & \textbf{98.63} \\
 \specialrule{.3em}{.2em}{.2em}
\end{tabular}
\end{table}

Application-Oriented License Plate (AOLP) dataset \cite{hsu2013application} has 2049 images of Taiwanese license plates, it is categorized into three subsets with different levels of difficulty for detection and recognition: access control (AC), traffic law enforcement (LE), and road patrol (RP). The evaluation protocol used in all previous work on AOLP is to train on two subsets and test on the third (e.g. for the RP results, we train on AC and LE and test on RP).

As shown in Table \ref{table:aolp}, we achieve state of the art results with small margin in AC and LE, and at big margin for the challenging RP subset. It is important to note that while \cite{wu2018many} uses extensive data augmentation on the shape and color of input license plate, we use only our described augmentation techniques and convert input image to grayscale (thus, going away with most color information).


\subsection{KHATT database}
\begin{table}[h!]
\caption{Results of our method compared to state of the art on the KHATT. Best results are in bold }
\label{table:kht}
\centering
\begin{tabular}{c c} 
 \specialrule{.3em}{.2em}{.2em}
 Method & CER(\%)\\
 \specialrule{.3em}{.2em}{.2em}
 HMM \cite{mahmoud2014khatt} & 53.87\% \\
 MDLSTM \cite{ahmad2017khatt} & 24.20\% \\
 Ours: 8(128,512)  & 9.9\% \\
 Ours: 16(128,512) & 9.5\% \\
 Ours: 16(128,1024) & \textbf{8.7\%} \\
 \specialrule{.3em}{.2em}{.2em}
\end{tabular}
\end{table}
KFUPM Handwritten Arabic TexT (KHATT) is a freely available offline handwritten text database of modern Arabic. It consists of scanned handwritten pages with different writers and text content. Pages segmentation into lines is also provided to allow the evaluation of line-based recognition systems directly without layout analysis. Forms were scanned at 200, 300, and 600 dpi resolutions. The database consists of 4000 paragraphs written by 1000 distinct writers across 18 countries. The 4000 paragraphs are divided into 2000 unique randomly selected paragraphs with different text contents, and 2000 fixed paragraphs with the same text content covering all the shapes of Arabic characters.

Following \cite{mahmoud2014khatt,ahmad2017khatt} we carry our experiments on the 2000 unique randomly selected paragraphs, using the database default split of these paragraphs into a training set with 4825 text-lines, test set with 966 text-lines, and a validation set with 937 text-line images.

The main preprocessing step we performed on KHATT was removing the large extra white spaces that is found on some line-images, as was done by \cite{ahmad2017khatt}. We had to perform this step as the alternative was using line-images with large height to account for this excessive space that can be, in some images, many times more than the line height.

As we do in all our experiments, we compare only with methods that report pure optical results (i.e. without using any language models or syntactic constrains). As shown in Table \ref{table:kht}, we achieve a significant reduction in CER compared to previous state-of-the-art, this is partly due to the complexity of Arabic script and partly due the fact that little work was previously done on KHATT.

\begin{table}[h!]
\caption{Comparison of our top performing methods on IAM. \textbf{C@N} is the CER@TopN(\%) metric.}
\label{table:iam2}
\centering
\begin{tabular}{c c c c c c c} 
 \specialrule{.3em}{.2em}{.2em}
 Method & C@1 & C@2 & C@3 & C@4 & C@5 & C@6 \\
 \specialrule{.3em}{.2em}{.2em}
 Ours: 8(128,1024)   & 5.8  & 5.1  & 4.8 & 4.6 & 4.4 & 4.3 \\ 
 Ours: 16(128,1024)  & 4.9  & 4.2  & 3.9 & 3.8 & 3.6 & 3.5 \\ 
 \specialrule{.3em}{.2em}{.2em}
\end{tabular}
\end{table}

\newcolumntype{C}[1]{>{\centering}m{#1}}

\subsection{IAM Handwriting Database}
\begin{table*}[t]
\caption{Results of our method compared to state of the art on the IAM Handwriting Database. \textbf{Aug.} is whether data augmentation is applied (True) or not (False). Best results are in bold }
\label{table:iam}
\centering
\begin{tabular}{C{4cm} C{4cm} c c c c } 
 \specialrule{.3em}{.2em}{.2em}
 Method & Preprocessing & Input Scale & Aug. & Val CER(\%) & Test CER(\%) \\
 \specialrule{.3em}{.2em}{.2em}
 5 layers MLP, 1024 units each \cite{bluche2015deep}  & De-skewing, De-slanting, contrast normalization, and region height normalization & 72 x W & No & - & 15.6  \\ 
 \hline
 7-layers BLSTM, 200 units each \cite{bluche2015deep} & Same as above & 72 x W & No & - & 7.3   \\ 
 \hline
 3-layers of [Conv. + 2D-LSTM], (4,20,100) units \cite{pham2014dropout} & mean and variance normalization of the pixel values & Original & No & 7.4 & 10.8   \\ 
 \hline
 3-layers of [Conv. + 2D-LSTM], wide layers \cite{castro2018icfhr2018} & De-slanting and inversion & Original & No & 5.35 & 7.85 \\  
 \hline
5-layers of [Conv. + 2D-LSTM], wide layers \cite{castro2018icfhr2018} & Same as above & Original & No & 4.42 & 6.64  \\  
 \hline
4-layers of [Conv. + 2D-LSTM], wide layers, double input conv. \cite{castro2018icfhr2018} & Same as above & Original & No & 4.31 & 6.39  \\  
 \hline
5 Convs. + 5 BLSTMs \cite{puigcerver2017multidimensional} & De-skewing and Binarization \cite{villegas2015modification} & 128 x W & No & 5.1 & 8.2   \\  
\hline
5 Convs. + 5 BLSTMs \cite{puigcerver2017multidimensional} & Same as above & 128 x W & Yes & 3.8 & 5.8   \\  
\hline
STN \cite{jaderberg2015spatial} + ReseNet18 + BLSTM stack \cite{duttaimproving} & De-skewing, De-slanting, Model pre-training & - & Yes & - & 5.7\\
\hline
Ours: 8(128,1024) & None & 32 x W & Yes & 4.1 & 5.8  \\  
\hline
Ours: 16(128,1024) & None & 32 x W & Yes & \textbf{3.3} & \textbf{4.9}  \\  
\specialrule{.3em}{.2em}{.2em}
\end{tabular}
\end{table*}

The IAM database \cite{marti2002iam} (modern English) is probably the most famous offline handwriting benchmark. It is compiled by the FKI-IAM Research Group. The dataset is handwritten by 657 different writers and composed of 1539 scanned text pages that correspond to English texts extracted from the LOB corpus \cite{johansson1980lob}. They are partitioned into writer-independent training, validation and test partitions of 6161, 966 and 2915 lines respectively. The original line images in the training set have an average width of 1751 pixels and an average height of 124 pixels. There are 79 different character classes in the dataset.

Since the primary concern of this work is image-based text recognition and how to get the best CER, purely out of visual data, we do not compare to any method that uses additional linguistic data to enhance the recognition rates of their systems (e.g. through the use of language models).

In Table \ref{table:iam} we give an overview and compare to pure optical methods in literature. As we can see, our method achieves an absolute 1\% decrease on CER (relative 16\% improvement) compared to previous state of the arts \cite{puigcerver2017multidimensional,duttaimproving} which used a CNN + BLSTM. Although we use input which is down-sampled 4 times compared to \cite{puigcerver2017multidimensional}, and although \cite{duttaimproving} (which achieves 0.1\% better CER than \cite{puigcerver2017multidimensional}) uses pre-training on a big synthetic dataset \cite{krishnan2016generating} and also uses test time data augmentation.

An important question in visual classification problems - and text recognition is no exception - is: given an ambiguous case, can our model disentangle this ambiguity and shortlist the possible output classes ? In a trial to answer this question we propose using the CER@Top$N$ metric which accepts $N$ output sequences for an input image and returns the minimum CER computed between the ground-truth and each of the $N$ output sequences. CTC beam search decoding can output a list of Top-N candidate output sequences, ordered by their likelihood as computed by the model. We use this list to compute CER@Top$N$. For all our experiments we use a constant beam width of 10. 

In Table \ref{table:iam2} we can see that when our model is allowed to give multiple answers it achieves significant gains, allowing a second choice gives a 15\% relative reduction in CER for both models. For our larger model, with only 5 possible output sequences (full lines), it can match the state of the art result on IAM \cite{voigtlaender2016handwriting} which uses a combination of word and character 10-gram language models, both trained on combined LOB \cite{johansson1986tagged}, Brown \cite{francis1971manual}, and Wellington \cite{bauer1993manual} corpora containing around 3.5M running words.

\subsection{ICFHR2018 Competition on Automated Text Recognition}
\begin{table*}[h!]
\caption{Results of our method compared to other methods submitted to ICFHR2018 Competition on Automated Text Recognition. Here we show the CER achieved by the system for each respective number of additional pages averaged over all 5 test documents. We also show the average CER for all submission for each of the 5 test documents. \textbf{Imp} is the ratio of CER at 16 pages to CER at 0 pages. Other methods results are taken from \cite{strauss2018icfhr2018}. Best results are in bold. }
\label{table:ic1}
\centering
\begin{tabular}{c|c c c c|c|c c c c c|c}
 \specialrule{.3em}{.2em}{.2em}
 \multirow{2}{*}{Method} & \multicolumn{4}{c|}{CER per additional specific training pages} & \multirow{2}{*}{Imp.} & \multicolumn{5}{c|}{CER per test document} & \multirow{2}{*}{total error} \\
 & 0 & 1 & 4 & 16 & & Konzil C & Schiller & Ricordi & Patzig & Schwerin \\
 \specialrule{.3em}{.2em}{.2em}
 Ours: 16(128,512) & \textbf{25.3474} & 12.6283 & \textbf{8.2786}  & \textbf{5.8248} & \textbf{22.9} & 6.4899  & \textbf{13.7660} & \textbf{17.3295} & \textbf{14.8451} & \textbf{12.3286} & \textbf{13.0198} \\
 Ours: 8(128,512)  & 25.9067 & \textbf{12.5891} & 8.3709  & 6.8175 & 26.3 & \textbf{6.4113}  & 14.1694 & 17.7747 & 15.0185 & 13.2164 & 13.4210 \\
 OSU               & 31.3987 & 17.7344 & 13.2672 & 9.0238  & 28.7 & 9.3941  & 21.0974 & 23.2664 & 23.1713 & 12.9845 & 17.8560  \\ 
 ParisTech         & 32.2516 & 19.7981 & 16.9794 & 14.7213 & 45.6 & 10.4938 & 19.0465 & 35.5964 & 23.8308 & 17.0202 & 20.9376  \\
 LITIS             & 35.2940 & 22.5078 & 16.8871 & 11.3448 & 32.1 & 9.1394  & 25.6926 & 30.5006 & 25.1841 & 18.0407 & 21.5084  \\
 PRHLT             & 32.7927 & 22.1470 & 17.8952 & 13.3288 & 40.6 & 8.6511  & 18.3928 & 35.0685 & 26.2566 & 18.6527 & 21.5409 \\
 RPPDI             & 30.8045 & 28.4038 & 27.2462 & 22.8461 & 74.1 & 11.9013 & 21.8799 & 37.2920 & 32.7516 & 28.5525 & 27.3252 \\
  
 \specialrule{.3em}{.2em}{.2em}
\end{tabular}
\end{table*}

\begin{table*}[h!]
\caption{CER per document when using 16 additional specific training pages for methods submitted to ICFHR2018 Competition on Automated Text Recognition. We also describe the architecture used by every method and whether they use data augmentation on training and testing or not. \textbf{<10} is the number of documents on which the CER is less than 10\%. Best results are in bold.}
\label{table:ic2}
\centering
\begin{tabular}{c c c c c c | c | C{3cm} C{2cm} c} 
 \specialrule{.3em}{.2em}{.2em}
 Method & Konzil C & Schiller & Ricordi & Patzig & Schwerin & < 10 & Architecture & Training Aug. & Test Aug. \\
 \specialrule{.3em}{.2em}{.2em}
 Ours   & \textbf{2.8278} & \textbf{8.1746}  & \textbf{11.4441} & \textbf{6.72669} & \textbf{2.2833} & \textbf{4} & 16(128,512) & Yes & No  \\ 
 \hline
 Ours   & 2.9467 & 8.5148  & 14.0927 & 6.77966 & 4.1130 & \textbf{4} & 8(128,1024) & Yes & No \\ 
 \hline
 OSU                 & 3.7874 & 12.4514 & 15.0412 & 12.5371 & 3.4996 & 2 & 7 Convs.  + 2 BLSTMs & Yes (grid distortion \cite{wigington2017data}, rescaling, rotation and shearing) & \makecell{Yes\\(37 copies)} \\
 \hline
 ParisTech           & 8.0163 & 14.5801 & 30.1986 & 15.5085 & 9.1846 & 2 & 13 Convs. + 3 BLSTMs + word LM & Yse \cite{chammas2018handwriting} & Yes \cite{chammas2018handwriting} \\
 \hline
 LITIS               & 4.8064 & 19.5665 & 16.3744 & 12.8284 & 6.6127 & 2 & 11 Convs. + 2 BLSTMs + 9-gram LM & Yes (baseline biasing) & No \\
 \hline
 PRHLT               & 4.9847 & 12.5486 & 28.5164 & 16.3506 & 7.1178 & 2 & 4 Convolutional Recurrent + 4 BLSTMs + 7-gram char-LM & No & No \\
 \hline
 RPPDI               & 9.1797 & 16.2908 & 30.4939 & 28.2998 & 24.9046 & 1 & 4-layer of [Conv. + 2D-LSTM], double input conv. \cite{castro2018icfhr2018} +4-gram word-LM & No  & No \\
 \specialrule{.3em}{.2em}{.2em}
\end{tabular}
\end{table*}

The purpose of this open competition is to achieve a minimum character error rate on a previously unknown text corpus from which only a few pages are available for adjusting an already pre-trained recognition engine \cite{strauss2018icfhr2018}. This is a very realistic setup and is much more probable than the usual assumption that we already have thousands of hand-annotated lines which is usually unrealistic due to the associated cost.

The dataset associated with the competition consists of 22 heterogeneous documents. Each of them was written by only one writer but in different time periods in Italian, and modern and medieval German. The training data is divided into a general set (17 documents) and a document-specific set (5 documents) of equal scripts as in the test set. The training set comprises roughly 25 pages per document. The test set contains the same 5 documents as the document-specific training set but uses a different set of pages: 15 pages for each document.

Each participant has to submit 4 transcriptions per test set based on 0, 1, 4 or 16 additional (specific) training pages, where 0 pages correspond to a baseline system without additional training. To identify the winner of the competition, CER of all four transcriptions for each of the 5 test documents is calculated (i.e. each submission is comprised of 20 separate files). Our system is currently the best performing system in this open competition achieving more than 25\% relative decrease in CER compared to the second place entry, as shown in competition website \cite{ICFHR2018comp}.

In Table \ref{table:ic1} we can see that our method achieves significant reduction in CER compared to other submissions reported in \cite{strauss2018icfhr2018}. We achieve the best result in every single metric reported in \cite{strauss2018icfhr2018}. Not only do our method achieve the best results in the various forms of CER reported in Table \ref{table:ic1}, but we also achieve the best improvement in CER from 0 specific training pages to 16 specific training pages with nearly 80\% decrease in CER which means our method is the most data-efficient of all submissions.

An important measure of method's quality is its applicability to real-life use case scenarios and resource constrains, so given well-transcribed 16 pages out of given document, how far can we go? This is shown in Table \ref{table:ic2}. To asses the significance of presented results, we use an empirical rule developed within the READ project \cite{EU_READ} (mentioned in \cite{strauss2018icfhr2018}), stating that transcriptions which achieve a CER below 10\% are already helpful as an initial transcription which can be corrected quickly by a human operator. We can see that while most submitted methods can manage at most 2 documents with CER less than 10\% when using 16 additional pages, our methods raises this number to 4 out of 5 tested documents. For the fifth document we achieve a CER of 11.4\% (slightly higher than 10\%). It is worth noting that while all training documents and the 4 other testing documents are German (from various dialects and centuries), this document (Ricordi) is Italian, and it is the only Italian resource in the whole dataset.

\subsection{Ablation Study}

\begin{table}[h!]
\caption{Results of using different gating functions in our ablation study. \textbf{CER} is the validation CER of the model. \textbf{CTC} is the validation CTC loss achieved by the model. Best results are in bold.}
\label{table:abs1}
\centering
\begin{tabular}{c l c c } 
 \specialrule{.3em}{.2em}{.2em}
 id & Method & CER(\%) & CTC\\
 \specialrule{.3em}{.2em}{.2em}
 1 & Baseline & 22.85 & 33.65 \\
 2 & $y_{i+1} = [ T(y_i) + 1 ] \cdot y_i$      & 27.82 & 41.64 \\
 3 & $y_{i+1} = H_1(y_i) \cdot T(y_i) + y_i$ & \textbf{22.46} & \textbf{33.17} \\
 4 & $y_{i+1} = [ T(y_i) + 1 ] \cdot H_1(y_i) - H_2(y_i) + y_i$ & 24.11 & 35.66 \\
 5 & $y_{i+1} = H_1(y_i) \cdot T(y_i) - H_2(y_i) + y_i$ & 25.65 & 37.81 \\
 6 & with residual connections, no gates & 25.53 & 37.30 \\
 7 & no residual connections, with gates   & 29.85 & 43.06 \\
 8 & no residual connections, no gates & 25.89 & 38.23 \\
 \specialrule{.3em}{.2em}{.2em}
\end{tabular}
\end{table}

\begin{table}[h!]
\caption{Results of using different normalization schemes in our ablation study. \textbf{CER} is the validation CER of the model. \textbf{CTC} is the validation CTC loss achieved by the model. Best results are in bold.}
\label{table:abs2}
\centering
\begin{tabular}{l c c } 
 \specialrule{.3em}{.2em}{.2em}
 Method & CER(\%) & CTC\\
 \specialrule{.3em}{.2em}{.2em}
 Baseline & \textbf{22.85} & \textbf{33.65} \\
 no Layer Normalization & 44.92 & 68.24 \\
 no Layer Normalization at start and end  & 29.29 & 42.93 \\
 no Batch Normalization & 35.59 & 50.81 \\
 no Softmax before GateBlocks & 24.52 & 36.12 \\
 Tanh instead of Softmax before GateBlocks & 24.43 & 36.11 \\
 \specialrule{.3em}{.2em}{.2em}
\end{tabular}
\end{table}

We here conduct an extensive ablation study to highlight which parts of our model are the most crucial on system's final classification performance. All experiments on this section are performed on the IAM handwriting dataset \cite{marti2002iam}, as it presents sufficient complexity to highlight the significance of different components of our system. We resize input images to 32 pixels height while maintaining the aspect ratio, however, we limit image width to 200 pixels and down-sample feature maps to a maximum of 4 x 50 pixels to limit the computational requirements needed for our experiments. We also use a small baseline model - 8(32,64) - for the same purpose. All experimented models use a fixed parameter budget of roughly 71k parameters.

\subsubsection{Gating Function}
First, we experiment with different gating functions other than the used baseline in Equation \ref{eqn:ctres}. For notational simplification we remove $P_1$ and $P_2$ from the equation, such that it becomes
\begin{equation}
y_{i+1} = \big[H_1(y_i) - H_2(y_i) \big] \cdot T(y_i) + y_i
\label{eqn:ctres2}
\end{equation}

In Table \ref{table:abs1}, the second model uses the gating function utilized by most previous work that made use of inter-layer gating functions in CNNs like \cite{wangfei2017residual,liao2018scene,gao2017reading,oktay2018attention}. It can be seen that our baseline model achieves a 20\% relative reduction in CER. This superiority may, however, be tied to our architectural choices. The third model uses a simplification of our proposed gating function with one transformation function, resembling the T-Only Highway network from \cite{greff2016highway}. It achieves slightly better CER than the baseline. We opted for the baseline as it gives roughly the same performance but with two transformation functions so it has more expressive power. The fourth and fifth models represent other variations on our proposed scheme where we try to change the input filtering order by performing multiplication before subtraction. The last three models illustrate the importance of gating and residual connections. It is interesting to note that using gates without residual connections is much worse than removing both of them.

\subsubsection{Normalization}

In Table \ref{table:abs2}, we show the effect of removing different parts of the normalization mechanisms utilized by our model. The second row shows that removing Layer Normalization has the most drastic effect on the model's performance. The third row shows that even just removing the two Layer Normalization layers at the beginning and end of the model increases CER by almost 30\%. The third shows that Batch Normalization is also crucial for model's performance, although not as important as Layer Normalization. The last two rows show the effect of removing Softmax normalization before GateBlocks or replacing it with just a Tanh activation function.

\section{Conclusion}
In this paper, we tackled the problem of general, unconstrained text recognition. We presented a simple data and computation efficient neural network architecture that can be trained end-to-end on variable-sized images using variable-sized line level transcriptions. We conducted an extensive set of experiments on seven public benchmark datasets covering a wide range of text recognition sub-tasks, and demonstrated state of the art performance on each one of them using the same architecture and with minimal change in hyper-parameters. The experiments demonstrated both the generality of our architecture, and the simplicity of adopting it to any new text recognition task. We also conducted an extensive ablation study on our model that highlighted the importance of each of its submodules.

Our presented architecture is general enough for application on many sequence prediction problem, a further research direction that we think is worthy of investigation is using it for automated speech recognition (ASR). Especially given its sample efficiency and high noise robustness as demonstrated by its performance on handwriting recognition and scene text recognition.

\ifCLASSOPTIONcompsoc
  \section*{Acknowledgments}
\else
  \section*{Acknowledgment}
\fi
We gratefully acknowledge the support of NVIDIA Corporation with the donation of the Titan X Pascal GPU used for this research.

\ifCLASSOPTIONcaptionsoff
  \newpage
\fi


\bibliographystyle{IEEEtran}
\bibliography{IEEEabrv,csr}

\end{document}